\documentclass[conference]{IEEEtran}
\IEEEoverridecommandlockouts
\usepackage{cite}
\usepackage{amsmath,amssymb,amsfonts}
\usepackage{algorithmic}
\usepackage{graphicx}
\usepackage{textcomp}
\usepackage{xcolor}
\usepackage{multirow}
\usepackage{booktabs}
\usepackage{makecell}
\usepackage{array}
\def\BibTeX{{\rm B\kern-.05em{\sc i\kern-.025em b}\kern-.08em
    T\kern-.1667em\lower.7ex\hbox{E}\kern-.125emX}}
\begin{document}

\title{Causal Coupled Mechanisms: A Control Method with Cooperation and Competition for Complex System}

\author{\IEEEauthorblockN{1\textsuperscript{st} Xuehui Yu}
\IEEEauthorblockA{\textit{Faculty of Computing} \\
\textit{Harbin Institute of Technology}\\
Harbin, China \\
yuxuehui0302@gmail.com}
\and
\IEEEauthorblockN{2\textsuperscript{nd} Jingchi Jiang}
\IEEEauthorblockA{\textit{the Artificial Intelligence Institute} \\
\textit{Harbin Institute of Technology}\\
Harbin, China \\
jiangjingchi@hit.edu.cn}
\and
\IEEEauthorblockN{3\textsuperscript{rd} Xinmiao Yu}
\IEEEauthorblockA{\textit{Faculty of Computing} \\
\textit{Harbin Institute of Technology}\\
Harbin, China \\
1190201203@stu.hit.edu.cn}
\and
\IEEEauthorblockN{4\textsuperscript{th} Yi Guan*}
\IEEEauthorblockA{\textit{Faculty of Computing} \\
\textit{Harbin Institute of Technology}\\
Harbin, China \\
guanyi@hit.edu.cn}
\and
\IEEEauthorblockN{5\textsuperscript{th} Xue Li}
\IEEEauthorblockA{\textit{Faculty of Computing} \\
\textit{Harbin Institute of Technology}\\
Harbin, China \\
20s103245@stu.hit.edu.cn}
}

\maketitle

\begin{abstract}
Complex systems are ubiquitous in the real world and tend to have complicated and poorly understood dynamics. For their control issues, the challenge is to guarantee accuracy, robustness, and generalization in such bloated and troubled environments. Fortunately, a complex system can be divided into multiple modular structures that human cognition appears to exploit. Inspired by this cognition, a novel control method, Causal Coupled Mechanisms (CCMs), is proposed that explores the cooperation in division and competition in combination. Our method employs the theory of hierarchical reinforcement learning (HRL), in which 1) the high-level policy with competitive awareness divides the whole complex system into multiple functional mechanisms, and 2) the low-level policy finishes the control task of each mechanism. Specifically for cooperation, a cascade control module helps the series operation of CCMs, and a forward coupled reasoning module is used to recover the coupling information lost in the division process. On both synthetic systems and a real-world biological regulatory system, the CCM method achieves robust and state-of-the-art control results even with unpredictable random noise. Moreover, generalization results show that reusing prepared specialized CCMs helps to perform well in environments with different confounders and dynamics.
\end{abstract}

\begin{IEEEkeywords}
complex system control, causal reasoning, hierarchical reinforcement learning
\end{IEEEkeywords}

\section{Introduction \label{1}}

\begin{figure}[h]
  \centering
  \includegraphics[width=\linewidth]{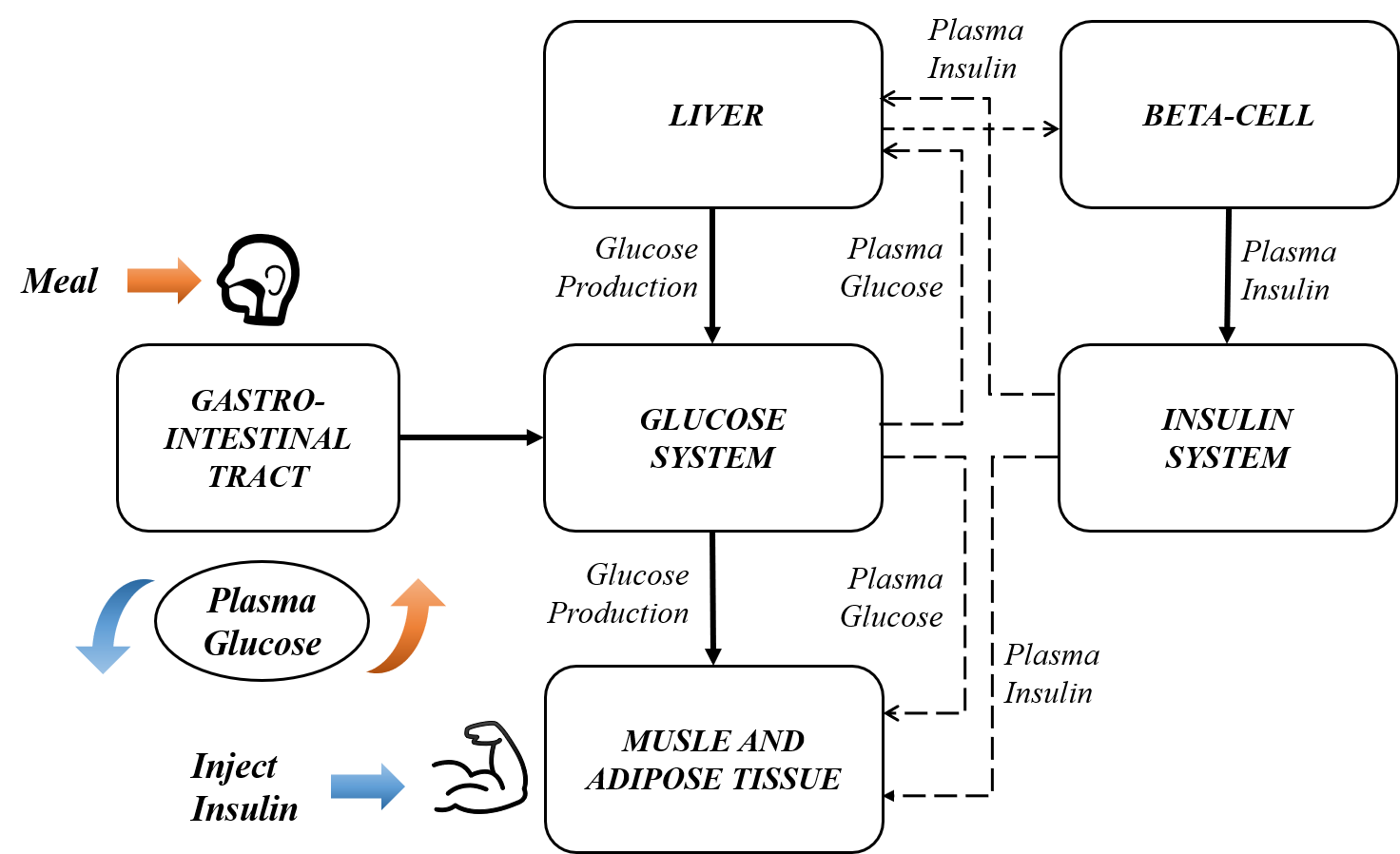}
  \caption{Modularization of a glucose-insulin control system. The glucose-insulin control system can be divided into insulin subsystem, glucose subsystem and other unit process models,and each needs to inform and influence each other.}
  \label{model}
\end{figure}

Control methods for complex systems are of critical importance \cite{chen2018optimal}. These complex systems represented by biological systems, transportation systems, and robotic systems have some common properties, such as the difficulty of understanding their dynamics and emergencies, which lead to a series of challenges to control them in our desired way \cite{baheti2011cyber}. By analysing many practical control tasks, we sum up three major challenges as follows:
\begin{itemize}
\item[*] $C1$: Computational complex dynamics, sometimes even with emergent properties and notably delays \cite{gupta2013transcriptional,glass2021nonlinear};
\item[*] $C2$: Unpredictable random external noises;
\item[*] $C3$: Heterogeneity among different objects. For example, different patient groups have different responses to treatment, yet the desired outcome is the same.
\end{itemize}
To solve the above three challenges for industrial application, improving the robustness and generalization of the control method is the core.

One important hallmark of human cognition is to exploit the modular structures of complex systems, which can be divided into multiple general and independent units \cite{goyal2019recurrent}. For example, the glucose-insulin control system can be divided into insulin subsystem, glucose subsystem and other unit process models as shown in Fig. \ref{model} \cite{dalla2007meal,dalla2007gim}. This cognition can help humen to better understand and control complex systems. The idea of dividing an entire system into independent functional mechanisms has also been studied in the field of causal reasoning, known as "causal modules". In causal reasoning, the independence between causal modules is a prerequisite for performing the division, and localized control \cite{pearl2000models}. However, causal boundary detection based on independence is intractable for high-dimensional variables. To simplify the detection of causal boundaries, an automatic division algorithm to rationalize the control tasks of each mechanism is necessary. Once agents learn the skills of controlling each mechanism by localized interventions, they can also control the entire system by combining and reusing these atomic skills. Reusing learned skills can solve transfer problems and achieve a stronger generalization.

After automatically dividing a complex system into multiple causal modules, how to concatenate these sub-modules to recover the original characteristics of the complex system needs to be considered. As phased by Aristotle, "the whole is greater than the sum of its parts;" properties of complex systems are not a simple summation of their independent functional mechanisms. Since the existing methods (modularity study \cite{goyal2019recurrent} and causal reasoning \cite{parascandolo2018learning,madan2021fast}) assume that only sparse interactions exist between the divided independent mechanisms, there is little attention to the cooperation between modules. However, the sparse interaction is a strong assumption that violates the laws of natural biological systems. As shown in Fig. \ref{model}, the liver can finish the degradation of insulin independently; the insulin flux which gets into the liver depends on the apportionment of the total insulin flux between plasma and liver. Plasma and the liver need to work together to maintain the entire system's stability. So agents must consider how to reuse the independent functional mechanisms and complete their information transfer in the second step.

In this paper, we propose a control method \textit{\textbf{Causal Coupled Mechanisms (CCMs)}} with state-of-the-art robustness and generalization, in which mechanisms contain full competition and cooperation. Competition and cooperation come from "The Origin of Species", which helps each natural biological community to complete the allocation of natural resources. In our modelling of a complex system, resources are the values of observable variables. An agent observes variables of systems and establishes control methods to assign them to each $CCM$ (i.e. automatic modularization).
Competition means that when an entire system is modularized, each $CCM$ competes for resources of the whole system to maintain their local functional integrity and maintain their steady state operation. In other words, competition makes each $CCM$ have the appropriate number of variables. The cooperation process considers how to reorganize the independent functional mechanisms. A hierarchical reinforcement learning (HRL) architecture is inspired by hierarchical processing information of human cognition. The high-level policy divides the complex system into multiple competing $CCMs$. The low-level policy focuses on each $CCM$ to maintain its regular operation. In cooperation processing, a cascade control module is used for hindsight and feedback regulation and a forward coupled reasoning (FCR) module recoveries helpful coupling information between variables and reduces computational complexity. 
We selected a real-world biological regulatory system, the glucose-insulin system \cite{dalla2007meal,dalla2007gim}, to verify the robustness and generalization of $CCM$ in control problems faced with the above three challenges. Generalization experiment is taken in environments with totally different confounders and dynamics. In addition, we design three validation experiments to verify the performance of $CCM$ on the three challenges respectively. The results show that $CCMs$ results in a diminished and dispersed effect of uncertainties and improves generalization. Finally, the visualization results of modularity show that different mechanisms have different functions.

\section{Related work \label{1.1}}
In recent years, many advanced methods have been used for the control of complex systems due to the decreasing cost of computation and sensors. Model predictive control has become a dominant control strategy in research on intelligent operation \cite{van2019comparison,bianchini2017integrated}. However such linear control models may not be accurate enough since the dynamics can be far from linear \cite{shaikh2014review}. Mainstream control method often use deep learning mainly in model-free end-to-end controller settings, such as control of cyber-physical systems \cite{wei2017deep,o2010residential}. And much of the success relies heavily on a reinforcement setup where the optimal state-action relationship can be learned via a large number of samples. To our knowledge, there are no high-performance model-free reinforcement learning (RL) models on biological control systems.

Recurrent Independent Mechanisms (RIMs) \cite{goyal2019recurrent} is an impetus for this paper. $RIMs$ is an architecture which divides complex systems into nearly independent transition dynamics, which communicate only sparingly through the bottleneck of attention. Such sparse interactions can reduce the difficulty of learning since few interactions need to be considered at a time, reducing unnecessary interference when a subsystem is adapted. Inspired by $RIMs$, we aim to divide the whole complex system into multiple functional mechanisms that have local independence. The key differences are that our $CCMs$ makes full communication between mechanisms to ensure that the system formed by reuse can restore the performance of the original system as much as possible, thereby improving the overall stability of the system. 

\begin{figure}[h]
  \centering
  \includegraphics[width=\linewidth]{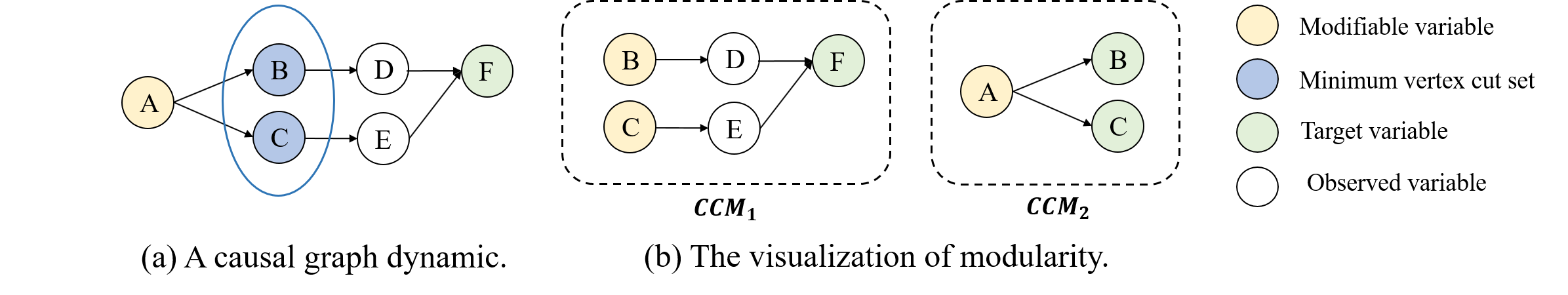}
  \caption{(a) An example of a CGD. The control task is to intervene in the modifiable variable $\{A\}$ and maintain the target variable $\{F\}$ in the goal range. (b) An example of dividing a CGD. Using $\{B, C\}$ as the cut, divide the CGD in Fig. \ref{causal_effect_reasoning} (a). $CCM_1$ is generated by severing the links to its parents and $CCM_2$ by severing the links to its children.}
  \label{causal_effect_reasoning}
\end{figure}

\begin{figure*}[!t]
  \centerline{\includegraphics[width=2\columnwidth]{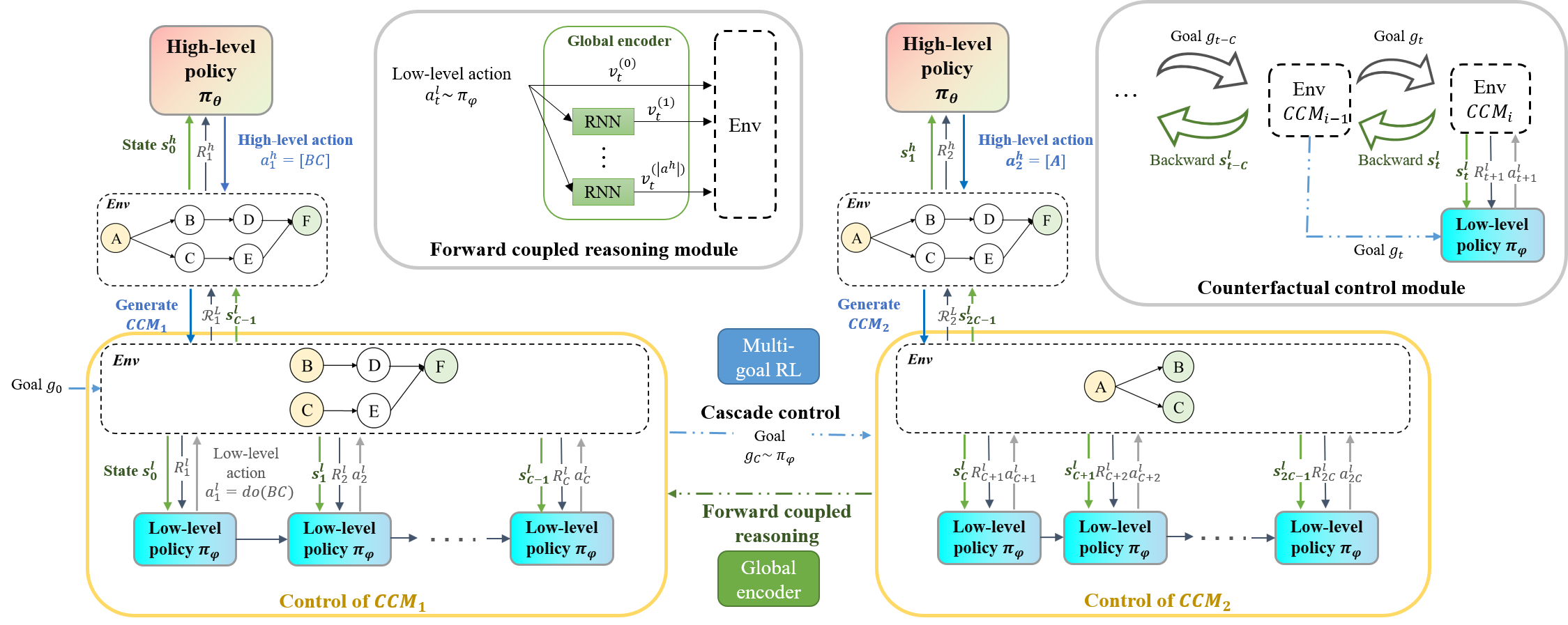}}
  \caption{Procedure for Causal Coupled Mechanisms. This figure presents the control process for the system in Fig. \ref{causal_effect_reasoning} (a). \textbf{Competition}: Firstly, the high-level policy $\pi_{\theta}$ divides the entire system into a subsystem $CCM$; Then the low-level policy $\pi_{\varphi}$ finishes the control of the $CCM$. These two steps are under the HRL framework. \textbf{Cooperation}: Time adjacent $CCMs$ need to contact under the help of two modules: a cascade control module helps the series operation of CCMs, and a forward coupled reasoning module is used to recover the coupling information lost in the division process.}
  \label{main_model_fig}
\end{figure*}

\section{Background \label{2}}
This section starts with the introduction to the environment: 1) the first is the modelling approach called "causal graph dynamic", and 2) the second is the components of environments. Then we introduce the related theories of HRL. 

The environment dynamic is defined as a causal graph dynamic:

\textbf{Causal Graph Dynamic (CGD).}
Agents interact with vertexes in the environment, leading to global dynamics. A causal dynamics are caused by neighbour-to-neighbour interactions and with a time-varying neighbourhood. Causal is not only used to express the cause-effect relationship (as one-way narrows showing in Fig. \ref{causal_effect_reasoning}) but also that the causes must lie within the past lightcone of the effect \cite{arrighi2012causal}. 

In a CGD, observable variables are divided into three types: modifiable, target and observed variables. The task of agents on CGDs is to keep the target variables of the system stable within the target range. Specifically, agents make actions, i.e. the control decision, to intervene in modifiable variables and lead to global dynamics. Finally, the effect of global dynamics is passed on to the target variable.

Inspired by the human hierarchical cognitive architecture, we adopt the standard continuous control HRL setting. 

\textbf{Hierarchical Reinforcement Learning.}
HRL algorithms automatically learn a set of primitive skills to help an agent accelerate learning. An HRL algorithm learns a low-level policy for performing each of the skills together with a high-level policy for sequencing these skills to complete the desired task \cite{nachum2018data}. Referencing the HIerarchical Reinforcement learning with Off-policy correction (HRIO) \cite{nachum2018data}, a two-level HRL approach that can learn off-policy, we chiefly improve the low-level policy so that primitive skills can cooperate. In addition, many detailed changes make HRIO more applicable to our tasks.

\section{Causal Coupled Mechanisms \label{3}}

The complex dynamical system can be divided into $K$ small subsystems (or mechanisms), i.e. an entire CGD is divided into $K$ small CGDs. We call such an independent and fully functional mechanism $CCM$ which has distinct functions that are unknown but can be sampled by taking actions. We propose the \textit{\textbf{Coupled Mechanisms}} assumption: the causal generative process of a system’s variables is composed of semi-autonomous modules that need to inform and influence each other. The significant difficulty is modularizing automatically.

The HRL architecture is used for automatic modularization of the whole system. Fig. \ref{main_model_fig} presents the process of $CCM$ for the system in Fig. \ref{causal_effect_reasoning} (a). The high-level policy $\pi_{\theta}$ with competitive awareness divides the whole complex system into multiple functional modules. Specifically, $\pi_{\theta}$ gives a high-level action $a^h$ according to the observation $s^h$ of the whole system to separate a $CCM$. The $CCM$ will be used as the environment of the low-level policy $\pi_{\varphi}$. There are two works for low-level policy: 1) controlling the subsystem $CCM$, and 2) helping coordinate between $CCMs$. Another two modules are needed for the cooperation, which is cascade control and FCR modules (more details can be found in Section \ref{3.2}).

\subsection{Competition between CCMs \label{3.1}}
In the HRL framework, the high-level policy $\pi_{\theta}$ divides the whole system into multiple $CCMs$, in which competitive relationships should remain among $CCMs$. Firstly, $CCMs$ should have functional integrity, i.e. $CCMs$ compete to maintain their size and avoid excessive modularization where $CCMs$ are too small. Secondly, $CCMs$ should be independent, i.e. $CCMs$ should decouple with useless vertexes and avoid $CCMs$ too big to operate independently even single $CCM$ is dominant. The low-level policy $\pi_{\varphi}$ finish the control task on each $CCMs$.

\subsubsection{High-level Policy for generating CCMs \label{3.1.1}} The high-level policy $\pi_{\theta}$ instructs the low-level policy $\pi_{\varphi}$ via high-level actions $a^h$, which it samples a new every $C$ steps. The high-level policy $\pi_{\theta}$ optimises itself by evaluating the task effect and obtaining timely feedback from the low-level policy $\pi_{\varphi}$. The feedback is a single-step average reward $\mathcal{R}^{L}$ that the low-level policy $\pi_{\varphi}$ gains at $C$ steps. The high-level RL is formalized with the quadruple $(\mathcal{S}^h, \mathcal{A}^h, \mathcal{P}^h, \mathcal{R}^h)$, whose elements are elaborated below.

\textbf{Actions.}
Actions given by the high-level policy $\pi_{\theta}$ are vertex sets, which are used for the whole system modularization. The effective idea is to minimize the loss of information in the modular process. We limit the action $a^h$ of the high-level policy $\pi_{\theta}$ to a minimum vertex cut set between modifiable variables and target variables for this purpose. As shown in Fig. \ref{causal_effect_reasoning} (a), the minimum vertex cut set $\{B, C\}$ is chosen for modularization. Action space $\mathcal{A}^h$ is the set of the minimum vertex cut set of $A \to F$.

\textbf{States.} 
The state $s^h_t$ is the description of the system and its statistical properties. More specifically, the $i$th dimension $s^{h,(i)}_t$ describes the statistical property of the $i$th minimum vertex cut set of $A \to F$:
\begin{equation}
  s^{h,(i)}_t=[isCon,dis,num,...],
  \label{eq1}
\end{equation}
where $isCon$ signifies whether the $i$th minimum vertex cut set is controllable (i.e. contained in the controllable subsystem), which is 1 if it is controllable, otherwise it is 0. $dis$ expresses the distance between the $i$th minimum vertex cut set and the target variables; $num$ shows the number of vertexes in the $i$th set. Beyond these, $s^{h,(i)}_t$ can also contain some other statistical properties, such as in-degree and out-degree.

\textbf{Transition.} 
The transition $\mathcal{P}^h(s_{t+1}^h|a_{t}^h, s_{t}^h)$ is the state transition probability used to identify the probability distribution of the next state $s_{t+1}^h$, which is defined as a map function: $\mathcal{P}^h:\mathcal{S}^h \times \mathcal{A}^h \to \mathcal{S}^h$.

\textbf{Reward.} 
Three principles are presented to design the reward function: (1) The feedback information given by the corresponding low-level policy $\pi_{\varphi}$ of the current $CCM_t$, e.g. $\mathcal{R}^{L}_t=\underset{\pi_{\varphi}}{max}\mathbb{E}[\sum^C_{i=0}[\gamma^{i}R^L(s_{t+i}^l)]/C]$; (2) vertexes in $CCM_t$ are all uncontrollable, i.e. no cross between the $CCMs$; (3) The episode length shouldn't be too long, which can cause $CCMs$ too small.The high-level reward function $R^H_t$ at time $t$ can be defined as:
\begin{equation}
  R^H_t= \begin{cases} \alpha \mathcal{R}^{L}_t + m - n, & \text {if $isCon=0$} \\ \alpha \mathcal{R}^{L}_t - m - n, & \text{if $isCon=1$} \end{cases} ,
  \label{eq2}
\end{equation}
where $m$ and $n$ are positive real numbers and penalty items for offending against the principle (2) and (3) separately.

Finally, high-level action $a_t^h$ is used to generate $CCM_t$ at time step $t$. Referring to the idea of cause-effect reasoning, the links from set $a_t^h$ to its parents and from set $a_{t-C}^h$ to its children are removed. The remained subgraph is current $CCM$.

\subsubsection{Low-level Policy for controlling CCMs \label{3.1.2}} The low-level policy finishes the control task of a $CCM$ in $C$ steps and returns a single-step average reward $\mathcal{R}^{L}$. The control goal $g_t$ generated in the cooperation with other $CCMs$, which will detail in Section \ref{3.2}. In RL architecture, low-level RL is formalized with the quartuple $(\mathcal{S}^l, \mathcal{A}^l, \mathcal{P}^l, \mathcal{R}^l)$, whose elements are defined as:

\textbf{Actions.} 
Agent's action is to intervene modifiable variables of $CCMs$ to $a_t^l$ at time step $t$. In causal theory, intervention vertexes are the "$do$ operator". Low-level action space is the entire real space. 

\textbf{States.} 
The low-level state $s_t^l$ is the values of vertexs in $CCM_t$.

\textbf{Reward.} 
The mission objective is to maintain target variables in the range of goal $g_t$. The piece-wise reward function refers to the tip of entity-to-box distance \cite{ren2020query2box} that uses an empirical approach aimed at maximising the ratio within goal $g_t$. So the low-level reward function is written as:
\begin{equation}
  R_t^L = \omega - dist_{outside}(s^l_t;\boldsymbol{q}) - \upsilon \cdot dist_{inside}(s^l_t;\boldsymbol{q}),
  \label{eq3}
\end{equation}
where $\omega$ represents a fixed scalar margin, $0 < \upsilon < 1$ is a fixed scalar, $\boldsymbol{q}=[q_{min}, q_{max}]=[g_t-\epsilon, g_t+\epsilon] \in \mathbb{R}^{2d}$ is a query box; $dist_{outside}(s^l_t;\boldsymbol{q}) = \lVert Max(s^l_t-q_{max},0) + Max(q_{min},0) \rVert_1
$ and $dist_{inside}(s^l_t;\boldsymbol{q}) = \lVert Cen(\boldsymbol{q})-Min(q_{max},Max(q_{min},s^l_t)) \rVert_1$; $Cen(\boldsymbol{q})=(q_{max}+q_{min})/2$ is the central point of $q_{max}$ and $q_{min}$. If the BG level for the next state is in the target range, the agent will receive a positive reward. Otherwise, it will receive a negative reward.  

The optimisal low-level policy obtains maximal single-step average reward $\mathcal{R}^{L}_t=\underset{\pi_{\varphi}}{max}\mathbb{E}[\sum^C_{i=0}[\gamma^{i}R^L(s_{t+i}^l)]/C]$ and backwards it to the high-level policy.

\subsection{Cooperation between CCMs  \label{3.2}}
Only with full cooperation among $CCMs$ the whole system can secure stability. To adaptively schedule tasks and cooperate, a cascade control module and a FCR module are proposed.

\subsubsection{Cascade Control Module \label{3.2.1}}

The agent's goal is to complete the system's stability, which means that the target variable of CGD is within the goal. Since the whole system is divided into multiple subsystems, the sub-goals of multiple stages are a phased decomposition of the init goal. The challenging initial goal will eventually be achieved by guiding agents to achieve periodic goals gradually. Hindsight experience replay \cite{andrychowicz2017hindsight} has given us great inspiration for generating sub-goals. The regenerated goals are generated by a function $m:\mathcal{S}\to\mathcal{G}$, i.e. corresponding goal can be found for any state. Generating appropriate goals is a big challenge in multi-goal RL but is natural in the setting of our problem. That is because we cascade $CCMs$ with their neighbours in space and time, shown on the right-hand side of Fig. \ref{main_model_fig}. The goal of the previous stage can be sampled from the state of the later stage. For example, $CCM_2$ is cascaded with $CCM_1$, in which target variables ${B,C}$ are the modifiable variables in $CCM_1$. The goal of target variables ${B,C}$ can sample from $CCM_1$, i.e. sample from the optimised low-level policy $g_C \backsim \pi_{\varphi}(g_C|the\ state\ of\ CCM_1, g_0)$. More formally, low-level policy generates low-level action $a^l_{t+i}\  (1 \leq i \leq C)$ according to the state $s^l_{t+i-1}$ of $CCM_t$, i.e. $a^l_{t+i} \backsim \pi_{\varphi}(s^l_{t+i-1},g_{t+i-1})$,
where $g_{t+i-1}$ is sampled from $\pi_{\varphi}(g_{t+i-1}|the\ state\ of\ CCM_1, g_{t+i-C-1})$, $g_{i}\ (i \leq C)$ is the init goal used in $CCM_1$, moreover, the goal of the whole complex system. 


\subsubsection{Forward Coupled Reasoning Module \label{3.2.2}} 
Some links between vertex cut set and its parent vertexes are removed during the generation of $CCMs$. When the vertex cut set contains more than one vertexes (i.e. modifiable variables), two issues arise 1) coupled information between modifiable variables will be lost, influencing action making; 2) agents need to control multiple actions simultaneously. Several studies that are apparently related to the second issue, but are actually computational complex, including multi-agent \cite{zhang2021multi} and multi-head \cite{flet2019merl} RL. We applied a global encoder for the two issues (shown in the middle part of Fig. \ref{main_model_fig}). The global encoder learns the coupling relationship between modifiable variables and reconstructs the other multiple actions according to one action generated by the policy. More specifically, low-level action $a^l_{t+i}$ is the intervention value for the first vertex in $a^h_t$ at time step $t+i$. The $j$th intervention value $v^{(j)}_{t+i}$ is defined as:
\begin{equation}
  v^{(j)}_{t+i} = \begin{cases} a^l_{t+i}, & \text {if $j=0$} \\ RNN \left( v^{(0)}_{1:t+i}, v^{(j)}_{1:t+i-1} \right) , & \text{if $j\not=0$} \end{cases} ,
  \label{eq7}
\end{equation}
where $0 \leq j \leq |a^h|$, $|a^h|$ is number of vertexes in $a^h$, $0 \leq i \leq C$. Low-level action is $v^{(0)}_{t+i} = a^{l}_{t+i} \backsim \pi_{\varphi}(s^{l}_{t+i-1}, g_{t+i-1})$. $RNN_{\xi}$ is an artificial neural network with parameter $\xi$, i.e. recurrent neural network. It takes the history values in and generates the current $v^{(j)}_{t+i}$.

\subsection{Optimization and Training  \label{3.3}}
In this section, we discuss how to optimize our framework. The objective function of the low-level policy network is to maximize the expectation of accumulated rewards of hierarchical decisions, 
\begin{equation}
  J^{L}(\theta) = \mathbb{E}\sum^C_{i=0}\gamma^{i}R^L_{t+i}/C .
  \label{eq8}
\end{equation}
where $C$ is the maximal low-level episode length. The objective function of the high-level policy is,
\begin{equation}
  J^{H}(\varphi) = \mathbb{E}\sum^T_{i=0}\gamma^{i}R^H_{t+i}/T ,
  \label{eq9}
\end{equation}
where $T$ is the maximal high-level episode length. Besides, we use policy gradient methods \cite{sutton1999policy} to optimize both high-level and low-level policies, and the particularized loss function varies by different RL methods in the experiments. The objective function for the FCR module is cross-entropy,
\begin{equation}
  J^{FCR}(\xi) = -\sum^N_{i=1}\left( \bar{v}_{i}ln(v_{i}) + (\bar{v}_{i} - 1) ln(1-v_{i}) \right) ,
  \label{eq10}
\end{equation}
where $v_{i}$ is the output of the global encoder $RNN_{\xi}$, $\bar{v}_{i}$ is the truth value from environment, and $N$ is the batch size.

\begin{figure}[h]
  \centering
  \includegraphics[width=\linewidth]{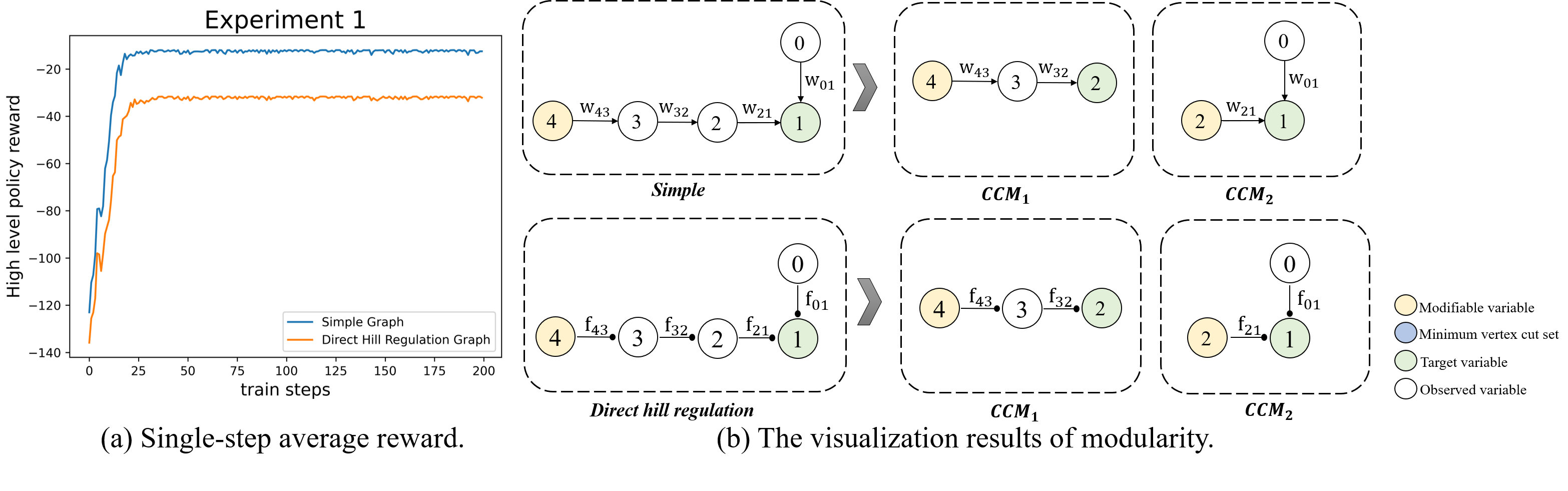}
  \caption{Experiment 1: (a) Average reward earned by the high-level agents in two environments. (b) The top left of the figure is causal graph for $Env\ 1$. A pointed arrow represents a linear relationship, and $w$ on the pointed arrow represents a constant. The down left of the figure is the causal graph for $Env\ 2$. We use a dotted arrow to represent a functional relationship, either activation or repression, with an implied explicit delay. The  independent functional $CCMs$ for $Env\ 1$ and $Env\ 2$ are in the top right-hand and bottom right-hand separately.}
  \label{high_level_policy_reward_1}
\end{figure}

\begin{figure}[h]
  \centering
  \includegraphics[width=\linewidth]{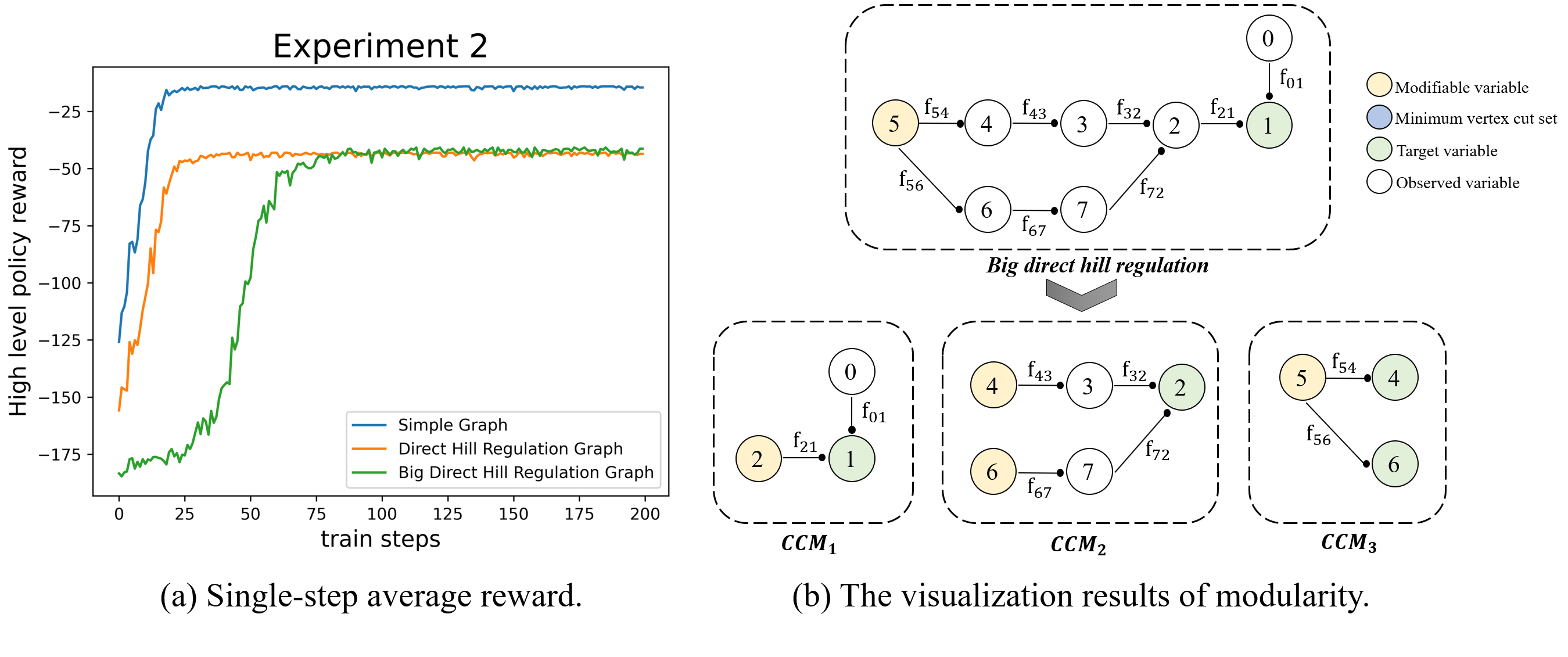}
  \caption{Experiment 2: (a) Average reward of high-level agents with external noises. (b) Top of the figure is the causal graph for $Env\ 3$. The more complex the system, the greater the number of $CCM$.}
  \label{high_level_policy_reward_2}
\end{figure}

\section{Experiments}
The main goal of our experiments is to show that the use of $CCMs$ improves robustness and generalization across changing environments and in modular tasks and to explore how it does so. We set up experiments for the three challenges mentioned above respectively. Experiment in Section \ref{4.1} corresponds to complex computations of challenge $C1$; Section \ref{4.2} corresponds to random external noise of $C2$; Section \ref{4.4} carries out generalisation verificatio which verifies $CCMs$ could deal with $C3$. Besides, an ordinary differential equation system \cite{dalla2007meal,dalla2007gim},which contains three challenges, is applied for robustness and module functional verification in Section \ref{4.3} and then used for generalisation verification in Section \ref{4.4}.

In $CCM$, we use the Advantage Actor-Critic (A2C) \cite{mnih2016asynchronous} to optimize both high-level and low-level policies and a Multi-layer perception (Mlp) to produce the policy, "CCM A2C-MlpPolicy" for short. As the baseline, we use several high-performance non-hierarchical RL algorithms: A2C-MlpPolicy (an A2C with Mlp policy); A2C-LstmPolicy (an A2C with long short-term memory policy); PPO-MlpPolicy (an proximal policy optimization \cite{schulman2017proximal} with Mlp policy).

\subsection{Experiment 1: Computational complex dynamics \label{4.1}}
We adopt two CGDs as environments and verify the impact of increased complexity (i.e. challenge $C1$). The maximum length of an episode is 100 for the two environments.

\begin{table*}
  \centering
  \caption{The single-step rewards for three experiments. All methods in Experiments 1 and 2 have learnt how to control the systems well, but the CCMs model obtained a higher reward. In the real-world system (Experiment 3), the baselines degraded so much that they could not complete the task, while the CCMs still obtained a positive reward.}
  \begin{tabular}{p{100pt}<{\centering}|p{60pt}<{\centering}|p{60pt}<{\centering}|p{60pt}<{\centering}|p{50pt}<{\centering}|p{50pt}<{\centering}|p{50pt}<{\centering}} 
\toprule
 \multicolumn{7}{l}{ \qquad  \qquad  \qquad  \qquad  \qquad  \qquad  \qquad  \qquad  \qquad \qquad  \qquad  \qquad  \qquad  \qquad \qquad \textbf{Experiment 1}}  \\ 
\hline
& \multirow{2}{*}{A2C-MlpPolicy} & \multirow{2}{*}{A2C-LstmPolicy} & \multirow{2}{*}{PPO-MlpPolicy} & \multicolumn{3}{l}{\qquad CCM A2C-MlpPolicy (ours)}  \\ 
\cline{5-7}
& &  &  & $CCM_1$ & $CCM_2$ & $CCM_2$ \\ 
\hline
Simple  & \textbf{24.00}  & 19.14 & 23.89 & \textbf{24.00} & 24.00 & -- \\
Direct Hill Regulation & 16.43 & 16.70 & 14.34 & \textbf{21.54} & 6.81 & -- \\
\toprule
\multicolumn{7}{l}{ \qquad  \qquad  \qquad  \qquad  \qquad  \qquad  \qquad  \qquad  \qquad \qquad  \qquad  \qquad  \qquad  \qquad \qquad \textbf{Experiment 2}}  \\ 
\hline
  & \multirow{2}{*}{A2C-MlpPolicy} & \multirow{2}{*}{A2C-LstmPolicy} & \multirow{2}{*}{PPO-MlpPolicy} & \multicolumn{3}{l}{ \qquad \qquad CCM A2C-MlpPolicy (ours)}  \\ 
  \cline{5-7}
  & &  &  & $CCM_1$ & $CCM_2$ & $CCM_3$\\ 
  \hline
  Simple  & 22.73  & 20.70 & 23.01 & \textbf{23.09} & 22.89 & -- \\
  Direct Hill Regulation & 7.14 & 8.19 & 6.91 & \textbf{9.36} & 7.58 & -- \\
  Big Direct Hill Regulation & 7.26 & 7.27 & 7.09 & \textbf{9.92} & 16.48 & 22.81 \\
  \toprule
\multicolumn{7}{l}{ \qquad  \qquad  \qquad  \qquad  \qquad  \qquad  \qquad  \qquad  \qquad \qquad  \qquad  \qquad  \qquad  \qquad \qquad \textbf{Experiment 3} }  \\ 
\hline
& \multirow{2}{*}{A2C-LstmPolicy} & \multirow{2}{*}{PPO-MlpPolicy} & \multicolumn{4}{l}{$\qquad \qquad \qquad \qquad $CCM A2C-MlpPolicy (ours)}{\centering}  \\ 
\cline{4-7}
&  &  & $CCM_1$ & $CCM_2$ & $CCM_3$ & $CCM_4$\\ 
\hline
Insulin (No noise)  & -899.93  & -1027.29 & \textbf{6.52} & 15.71 & 31.66 & -- \\
Insulin (Random large noise) & -899.93 & -1027.29 & \textbf{6.52} & 3.98 & 16.12 & 22.07 \\
\toprule

  \end{tabular}
  
  \label{result2}
  \end{table*}

\textbf{Env 1: Simple.}
This CGD contains five variables. With \cite{dasgupta2019causal} as reference, each node $x_i\in \mathbb{R}$ is a Gaussian random variable. Parentless observed variables have distribution $\mathcal{N}(\mu=0.0, \delta=0.1)$. A node $x_i$ with parents $pa(x_i)$ has conditional distribution $p(x_i|pa(x_i))=\mathcal{N}(\mu=\sum_j w_{ij} x_j, \delta=0.1)$, where $x_j\in pa(x_j)$ and $w_{ij}$ is a constant. The graph structure is represented at the top left of Fig. \ref{high_level_policy_reward_1} (b).

\textbf{Env 2: Direct hill regulation.}
Physiological processes typically have emergent properties of many parameters or time delays between the initiation of the physiological mechanism and the resulting functional output \cite{d2008synchronization,glass2021nonlinear}. To model the phenomenology of many biological networks, we refer to explicit-delay modeling \cite{glass2021nonlinear} to build CGDs. Specifically, the conditional distribution $p(x_i|pa(x_i))=\mathcal{N}(\mu=\sum_j f_{ij} x_j, \delta=0.1)$ is changed, where $f_{ij}$ represents explicit-delay function (i.e. $f_{ij}$ is a direct hill regulation function in $Env\ 2$). The other situations keep the same, as shown at the top right of Fig. \ref{high_level_policy_reward_1} (b).

\textbf{Result.}
Fig. \ref{high_level_policy_reward_1} (a) shows the result of the high-level policy on modularity. The single-step average reward goes steadily up in the two environments, although it decreases with the increase of system complexity. From the visual results in Fig. \ref{high_level_policy_reward_1} (b), the whole system is divided into two parts and the vertex cut set $\{2\}$ is chosen. An obvious phenomenon can be found that the $CCM_2$ in Fig. \ref{high_level_policy_reward_1} (b) contains a collider $x_1$ (i.e. a nonendpoint vertex at where two arrowheads meet) which will increase complexity. The existence of collider $x_1$ is perhaps a good reason that agents choose to modularize in $x_2$. In the child table "Experiment 1" of table \ref{result2}, the column headed "$CCM_1$" represents the single-step reward for controlling the target variables of the whole system. $CCMs$ achieves the highest reward and is relatively robust to increasing the complexity of systems, whereas the baseline's performance degraded more. 

\subsection{Experiment 2: Unpredictable random external noises \label{4.2}}
Parentless target variables will randomly become a larger value (increased more than ten times) to verify robustness. Apart from causal graphs in $Experiment\ 2$, a new environment $Env\ 3$ is designed. The maximum length of an episode is 100 for the three environments.

\textbf{Env\ 3: Big Direct hill regulation.}
$Env\ 3$ adds some variables and a double path based on $Env\ 2$ to increase system complexity, which can better show the impact of external noises (as shown at the top of Fig. \ref{high_level_policy_reward_2} (b)). The double paths can help check whether the FCR module can restore coupled information between modifiable variables. 

\textbf{Result.}
Random external noise reduces the reward for both high-level policy and low-level policy, compared with "Experiment\ 1". The performance on $Env\ 2$ degrades more severely because of the addition of computational complexity and random external noise. When evaluating in the environment with novel unseen noise, $CCMs$ doesn’t achieve perfect performance but strongly outperforms the baselines (shown as child table "Experiment 2" of \ref{result2}). In the visualization results (Fig. \ref{high_level_policy_reward_2} (b)), $CCMs$ is consciously trying to divide the $Env\ 3$ into three parts to reduce complexity. The divisions for $Env\ 1$ and $Env\ 2$ is the same as Fig. \ref{high_level_policy_reward_1} (b) in "Experiment\ 1". This indicates that the agent will use different modularity strategies depending on the system's complexity.

The modularization of $CCMs$ results in a diminished and dispersed effect of the huge noise, enhancing the target information and suppressing irrelevant noise. As shown in Fig. \ref{high_level_policy_reward_2} (b), $x_0$ containing noise is divided into $CCM_1$ and has diminished effect on $CCM_2$. The agent has learned how to eliminate the influence of unseen noise on the system's stability in a smaller graph $CCM_2$.

\begin{figure}[h]
  \centering
  \includegraphics[width=\linewidth]{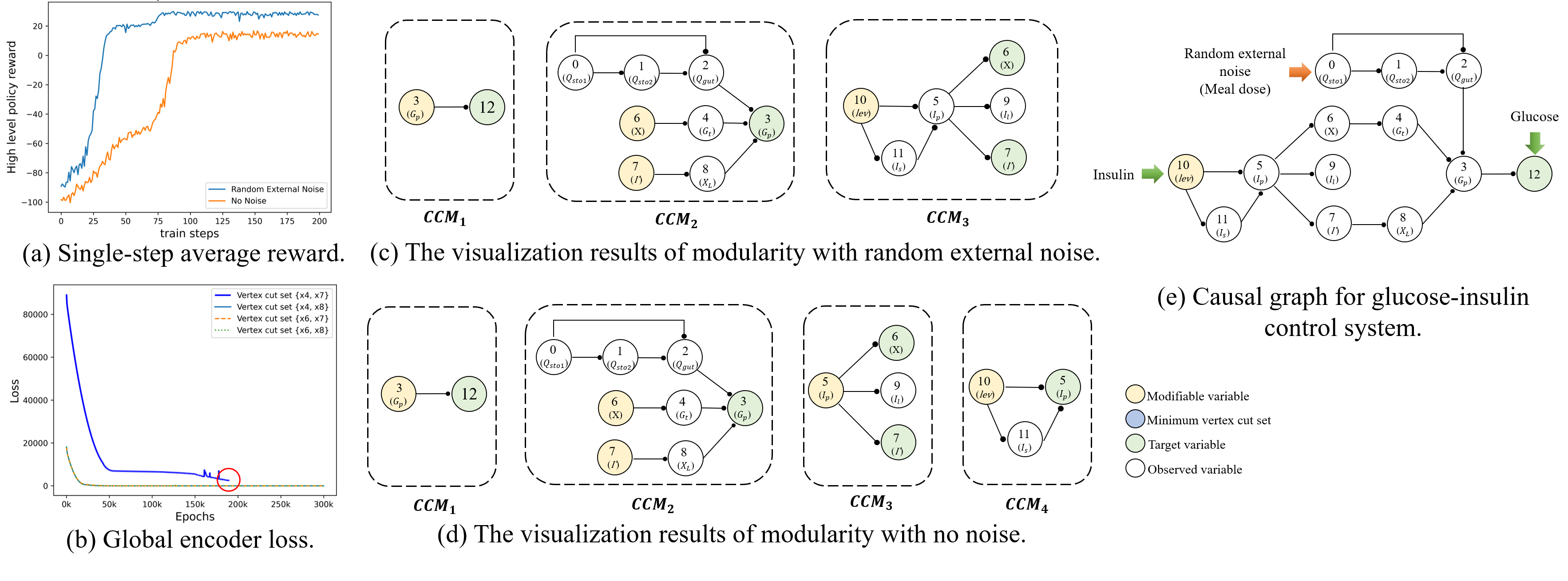}
  \caption{Experiment 3: The results of high-level policy for glucose-insulin system and The causal graph for glucose-insulin control system. 
  We use a dotted arrow to represent a functional relationship, and the specific function definition is the same as that in \cite{dalla2007meal,dalla2007gim}. }
  \label{high_level_policy_reward_3}
\end{figure}

\subsection{Experiment 3: A real-world biological regulatory system \label{4.3}}

\begin{table*}
  \caption{Generalization results in 30 individuals. The figures in the table are TIR values.}
  \centering
    \begin{tabular}{|p{30pt}<{\centering}|p{12pt}<{\centering}|p{16pt}<{\centering}|p{16pt}<{\centering}|p{16pt}<{\centering}|p{16pt}<{\centering}|p{20pt}<{\centering}|p{12pt}<{\centering}|p{16pt}<{\centering}|p{16pt}<{\centering}|p{16pt}<{\centering}|p{16pt}<{\centering}|p{20pt}<{\centering}|p{12pt}<{\centering}|p{16pt}<{\centering}|p{16pt}<{\centering}|p{16pt}<{\centering}|p{16pt}<{\centering}|}
    \toprule
    \multicolumn{2}{l}{} & CCM1 & CCM2 & CCM3 & CCM4 & \multicolumn{2}{l}{} & CCM1 & CCM2 & CCM3 & CCM4 & \multicolumn{2}{l}{} & CCM1 & CCM2 & CCM3 & CCM4 \\ 
    \hline
    \multirow{10}{*}{adolescent} & \#001       & 1.00   & 0.00   & 0.90   & 0.73    & \multirow{10}{*}{adult}      & \#001            & 1.00   & 0.38   & 0.88   & 0.66 &    \multirow{10}{*}{child}      & \#001            & 0.88   & 0.14   & 0.60   & 0.61   \\
                                 & \#002       & 0.98   & 0.30   & 0.88   & 0.65    &                                 & \#002            & 1.00   & 0.00   & 0.94   & 0.80 &                                 & \#002            & 1.00   & 0.07   & 0.56   & 0.73    \\
                                 & \#003       & 1.00   & 0.10   & 0.54   & 0.74    &                                 & \#003            & 1.00   & 0.26   & 0.84   & 0.69 &                                 & \#003            & 1.00   & 0.14   & 0.79   & 0.73    \\
                                 & \#004       & 1.00   & 0.13   & 0.87   & 0.70    &                                 & \#004            & 1.00   & 0.33   & 0.84   & 0.63  &                                 & \#004            & 1.00   & 0.12   & 0.42   & 0.57    \\
                                 & \#005       & 1.00   & 0.37   & 0.89   & 0.68    &                                 & \#005            & 1.00   & 0.31   & 0.92   & 0.77 &                                 & \#005            & 1.00   & 0.17   & 0.74   & 0.74    \\
                                 & \#006       & 1.00   & 0.18   & 0.30   & 0.54    &                                 & \#006            & 1.00   & 0.28   & 0.29   & 0.47&                                 & \#006            & 1.00   & 0.13   & 0.78   & 0.69    \\
                                 & \#007       & 1.00   & 0.10   & 0.90   & 0.66    &                                & \#007            & 1.00   & 0.14   & 0.32   & 0.52 &                                 & \#007            & 1.00   & 0.12   & 0.56   & 0.61    \\
                                 & \#008       & 1.00   & 0.17   & 0.92   & 0.74    &                                 & \#009            & 1.00   & 0.27   & 0.91   & 0.78 &                                 & \#008            & 0.67   & 0.09   & 0.87   & 0.66    \\
                                 & \#009       & 1.00   & 0.04   & 0.91   & 0.70    &                                 & \#008            & 1.00   & 0.00   & 0.75   & 0.66 &                                 & \#009            & 1.00   & 0.21   & 0.90   & 0.73    \\
                                 & \#010       & 1.00   & 0.32   & 0.76   & 0.60    &                                 & \#010            & 1.00   & 0.00   & 0.91   & 0.76 &                                 & \#010            & 1.00   & 0.14   & 0.80   & 0.65    \\ 
  \toprule
  \end{tabular}
  \label{app_2}
\end{table*}

We verify the performance of $CCMs$ on a real-world biological regulatory system \cite{dalla2007meal,dalla2007gim}, which is suitable for $CCMs$. The model assumes that the glucose and insulin subsystems are linked one to each other by the control of insulin on glucose utilization and endogenous production (the original gives more detailed divisions). We can compare the modularization results of $CCMs$ with the actual functional modules to investigate the functionality of automatic modularization. According to the definition of CGD, the graph structure is shown in Fig. \ref{high_level_policy_reward_3} (e). The glucose-insulin model has 26 free parameters that vary in different individuals and will be used in generalization experiment. The glucose-insulin meal model provides thirty individuals, and Adult \#004 is selected to interact with $CCMs$. Each step simulates one minute, and the maximum length of an episode is 1440, i.e., one natural day. In this experiment, noises are designed as two types, i.e. with and without noise. In the real world, the uncertainties (e.g. sensor error, irregular eating habits and so on) in glucose control of diabetics can be linked to the random external noise.

\textbf{Result.}
The results of high-level policy are shown in Fig. \ref{high_level_policy_reward_3}. In the child table "Experiment 3" of table \ref{result2}, baselines failed to control the system under the two noise settings while $CCMs$ achieved good performance. The loss curve of global encoders is displayed in Fig. \ref{high_level_policy_reward_3} (b), and the red circle represents the phenomenon of radiant explosion. In addition, we make a deeper analysis:

\begin{figure}[h]
  \centering
    \centerline{\includegraphics[width=\columnwidth]{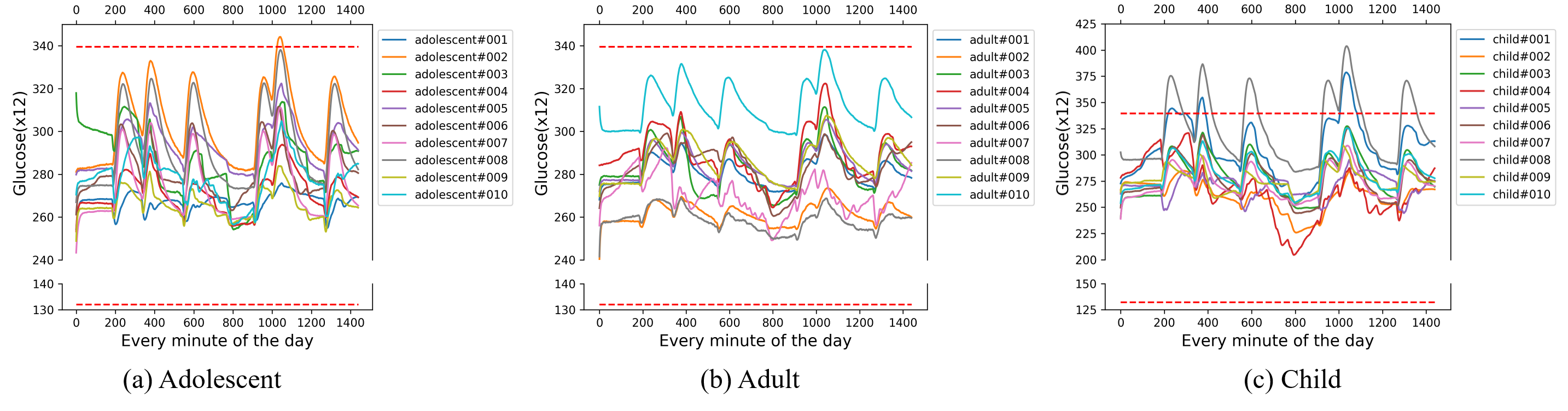}}
    \caption{The glucose profile over an optimised $CCMs$ trained with Adult \#004. 
    A subject with a consistent range above 180 mg/dl is generally held to have hyperglycemia, whereas a consistent range below 70 mg/dl is considered hypoglycemic. 70 and 180 multiplied by 1.8 correspond to the two red dashed lines.}
    \label{glucose}
  \end{figure}

\textit{(1) Analysis for multipath selection and the effect of FCR module.} At the second step, the high-level agent needs to choose a double vertexes set for modularization. In the experiment with noises, high-level agents only learnt the $x_{12} \overset{6.52}{\to} x_3 \overset{3.98}{\to} x_6x_7 \overset{16.12}{\to} x_5 \overset{22.07}{\to} x_{10}$ (the value on arrow represents the single-step average reward). In the noiseless experiment, high-level agents found three vertex cut sets $\{x_6, x_7\}$, $\{x_6, x_8\}$ and $\{x_4, x_8\}$, and chose the best one finally. The FCR for these three sets also convergence as shown in Fig. \ref{high_level_policy_reward_3} (b). In summary, 1) the agents weighed up all feasible control paths and chose the best, 2) the choice of control path is directly related to the effect of FCR and 3) the FCR can extract useful coupling information.

\textit{(2) Accuracy of realistic tasks.} We take testing for 1440 steps and calculate the percentage time in the glucose target range of [70, 180] mg/dL (TIR) \cite{maahs2016outcome}. In the training task adult \#004, the TIR is 100\%, which means that the policy given by $CCMs$ is of practical significance and can be used for glucose control tasks.

\textit{(3) Functionality of the mechanisms.}The four mechanisms have different functions. For example, $x_3 \to x_{12}$ in $CCM_1$ represents the dynamic relationship between plasma and interstitial fluid glucose.

\subsection{Experiment 4: Heterogeneity among different objects \label{4.4}}
The functional mechanisms in $CCMs$ have an ingenious connection with the concept of affordances from cognitive psychology \cite{cisek2010neural}. An agent should be ready to adapt immediately, even in an environment of uncertainty, executing skills which are at least partially prepared. This suggests that agents should contextually process sensory information, building representations of potential actions that the environment currently affords. We investigate how agents use prepared specialised $CCMs$ to improve generalisation between different environments, which have important variation factors. Adult \#004 interacts with $CCMs$ for training, and thirty individuals are used for testing. 

\textbf{Result.} \textit{(1) Analysis for generalization.}Similar to "Experiment\ 3", we take testing for 1440 steps and calculate the TIR for 30 individuals, which is shown in the column headed "$CCM_1$ in the table \ref{app_2}. The agent can adjust immediately to varied environments executing skills which are at least partially prepared.

\textit{(2) Visualization of realistic tasks.}The glucose profile over 1440 steps is employed for demonstration purposes over a testing period. Among 1440 steps, we apply three random noise interventions to the individuals. Fig. \ref{glucose} shows the change of glucose under the control of the optimised $CCMs$. The value of glucose is always within the normal range in the adult group. Hyperglycemia occurred in Adolescent \#002, Child \#001 and Child \#008, suggesting the child group differs more significantly from the adult group. In summary, $CCM$ has perfect generalisation in the same population and certain generalisation in different populations.

\section{Conclusion}
In this paper, we propose a hierarchical reinforcement learning method to control complex systems. This method incorporates the high-level agent with competition awareness and the low-level agent with cooperation awareness to search for the optimal control policy. To improve the robustness and generalization of our control algorithm, Causal Coupled Mechanisms, the high-level policy employs multi-goal learning to divide an entire system into multiple atomic causal modules, which can be effectively transferred and reused in similar but different tasks. To relax the independent assumption between causal modules, the low-level policy adopts coupled reasoning to deal with cooperative control problems. We conduct experiments on both synthetic systems and a real-world biological regulatory system to verify the advantages of our model. Compared to the existing methods, our method achieves the state-of-the-art results in addressing the complexity, random external noise, and heterogeneity of complex systems.

\section{Acknowledgement}

This study was supported in part by a grant from National Natural Science Foundation of China [62006063] and the Heilongjiang Provincial Postdoctoral Science Foundation (CN) [LBH-Z20015].

\small
\bibliographystyle{IEEEtran}
\bibliography{my_bib}
\end{document}